\documentclass{article}

\usepackage{subcaption}
\usepackage{microtype}
\usepackage{graphicx}
\usepackage{graphics}
\usepackage{wrapfig}
\usepackage{float}
\usepackage{caption}
\usepackage{booktabs} 
\usepackage{rotating}

\usepackage{hyperref}

\usepackage[preprint]{neurips_2022}
\usepackage{amsmath}
\usepackage{amssymb}
\usepackage{mathtools}
\usepackage{amsthm}

\usepackage[capitalize,noabbrev]{cleveref}

\theoremstyle{plain}
\newtheorem{theorem}{Theorem}[section]
\newtheorem{proposition}[theorem]{Proposition}

\theoremstyle{definition}
\newtheorem{definition}[theorem]{Definition}

\theoremstyle{remark}

\usepackage[textsize=tiny]{todonotes}

\usepackage{algorithm}
\usepackage{algpseudocode}

\title{Out-of-Distribution Detection using BiGAN and MDL}

\author{%
  Mojtaba Abolfazli \textsuperscript{1}\\
  \texttt{mojtaba@hawaii.edu} \\
   \And
   Mohammad Zaeri Amirani \textsuperscript{1}\\
   \texttt{zaeri@hawaii.edu} \\
   \AND
   Anders H{\o}st-Madsen \textsuperscript{1}\\
   \texttt{ahm@hawaii.edu} \\
   \And
   June Zhang \textsuperscript{1} \\
   \texttt{zjz@hawaii.edu} \\
   \And
   Andras Bratincsak \textsuperscript{2} \\
   \texttt{andrasb@hphmg.org} \\
   \\
   \textsuperscript{1} Department of Electrical and Computer Engineering\\
   \textsuperscript{2} Department of Pediatrics, John A. Burns School of Medicine \\
   University of Hawaii at Manoa, Honolulu, HI, USA
}

\begin{document}

\maketitle

\begin{abstract}
We consider the following problem: we have a large dataset of
normal data available. We are now given a new, possibly quite
small, set of data, and we are to decide if these are normal data,
or if they are indicating a new phenomenon. This is a novelty
detection or out-of-distribution detection problem. An example is in medicine,
where the normal data is for people with no known disease, and the
new dataset people with symptoms. Other examples could be in security.
We solve this problem by training a bidirectional generative adversarial network (BiGAN) on the normal data and
using a Gaussian graphical model to model the output. We then use
universal source coding, or minimum description length (MDL) on the output to decide if it is a new
distribution, in an implementation
of Kolmogorov and Martin-L\"{o}f randomness.
We apply the methodology to both MNIST data and a real-world electrocardiogram (ECG) dataset of healthy and patients with Kawasaki disease,
and show better performance in terms of the ROC curve than similar
methods.
\end{abstract}

\section{Introduction}
The basic problem we consider is the following:
we have a large dataset $\mathbf{x}_i$, $i=1\ldots N$ of `normal' data
available for training. For example, this could be ECG data for people with no known
heart disease. 
We are then given
a test set $\mathbf{x}_i'$, $i=1\ldots M$ with $M\ll N$
and wants to determine if this belongs
to the class of normal data, or if this should be considered
a new class of data. If $M=1$ this is
anomaly detection or outlier detection; if
$M>1$ this can be called novelty detection, based
on the terminology in \cite{Pimentel14} or group anomaly detection,
as in \cite{chalapathy2018group}.

Alternatively, if we consider class identity to be solely based on the data distribution, then we can view novelty detection as an out-of-distribution (OOD) 
or distribution shift test based on \cite{NalisnickAl19,RabanserAl19}.

When $M=1$, the only real possibility of detecting anomalous data is to check if it fits with normal data,
i.e., if it was likely to have been generated by the distribution
underlying the normal data. When $M>1$, more approaches to analysis open up.
One can still check the likelihood of the test data under the distribution of the normal data. But since we have more test data samples available, one can try to find
new patterns in data that could be an expression of
a new underlying model.
This can be expressed as finding an alternative
distribution for the test data.

The goal of our paper is to perform OOD detection on high-dimensional data, $\mathbf{x}_i'$, $i=1\ldots M >1$, particularly when there are only subtle differences between the test and training data. Our method uses Rissanen's minimum description length (MDL) \cite{Rissanen78,Rissanen83}, which explicitly accounts for the complexity of the probabilistic model to avoid overfitting.

We focus on subtle changes as these are typical
of for example adversarial attacks or medical
data. Adversarial attackers want to mimic
normal data to go undetected. And in for example
ECG data, many heart conditions are difficult to
detect in a standard 12-lead electrocardiogram (ECG). The indications
is in subtle changes of relationship between multiple
variables.

\section{Related Work}
\label{sec:RelatedWork}
The problem of novelty or OOD detection using deep generative models has attracted much attention in recent years due to the ability of these networks to estimate the distribution of training data. The central approach to this problem is that these models give lower likelihood to OOD data than in-distribution data. This approach was initially suggested in \cite{bishop1994novelty}. 
However, further investigations in recent years show that generative models may give higher likelihood to OOD data than in-distribution data in high dimensional cases \cite{choi2018waic, nalisnick2018deep}. This phenomenon makes their applicability for novelty or OOD detection limited.
\cite{choi2018waic} use an ensembles of generative models to tackle this issue. In \cite{NalisnickAl19}, a method based on typicality set in information theory was suggested to detect OOD data based on the closeness of the information content to the empirical entropy of training data. In a similar work \cite{morningstar2021density}, multiple statistics from in-distribution data are extracted to differentiate between in-distribution and OOD data. A parameter-free OOD score similar to likelihood-ratio was proposed in \cite{serra2019input}. A combination of Rao's score test \cite{rao1948large} and typicality test \cite{NalisnickAl19} was suggested in \cite{bergamin2022model}. \cite{chalapathy2018group} proposed an approach based on measuring the deviation from in-distribution data. 

All these approaches more or less are based on
how well data fits with the default distribution,
that is essentially based on the likelihood, eq. (\ref{eq:likelihood}) below. As explained after (\ref{eq:likelihood}) this has an inherent limitation.
Our approach is based on finding an alternative model
for the OOD data through MDL.

\section{Problem Statement}
\label{sec:ProbState}
Let the data be $\mathbf{x} \in\mathbb{R}^n$.
We model the normal data through
a probability distribution $P(\mathbf{x})$.
In an ideal setting, this model is given; but
in practical scenarios, the model is learned
from training data. How this is done is the main
topic of the paper. We will start by discussing
the problem for known $P(\mathbf{x}$) and then generalize
to unknown $P(\mathbf{x})$.

\subsection{$P(\mathbf{x})$ Known }
Given a set of test data $\mathbf{x}_i'$, $i=1\ldots M$,
the OOD problem can be stated as
the hypothesis test
\begin{align*}
  H_0&: \{\mathbf{x}_i'\}\sim P \nonumber\\
  H_1&: \{\mathbf{x}_i'\}\sim \widehat P
\end{align*}
with $\widehat P\neq P$, with $\widehat P$ \emph{not} known. One approach is
to check how well the data fits $P$ through the log-likelihood of $\{\mathbf{x}_i'\}$ under the (known) distribution $P$, 
\begin{equation}
    \ell(\{\mathbf{x}_i'\})=-\sum_{i=1}^M\log_2 P(\mathbf{x}_i'),\label{eq:likelihood}
\end{equation}
as was done in \cite{NalisnickAl19}. We consider base-2 logarithm instead of natural logarithm so that we can interpret the log-likelihood as binary codelength via Kraft's Inequality \cite{}.

However, it is possible that $\{\mathbf{x}_i'\}\sim \widehat P$ and  $\widehat P\neq P$, but
$\{\mathbf{x}_i'\}$ fits perfectly with $P$. One example
is that $P$ is iid across the $n$-components of $\mathbf{x}$, and $\widehat P$ has the same marginals as $P$ but a different
joint distribution. It is therefore necessary to have an alternative
distribution $\widehat P$, and since it is unknown, it has to be estimated
from data. One such test in one
dimension is the Kolmogorov-Smirnoff test \cite{corder2014nonparametric}, which compares
the CDF $F$ and the empirical CDF of $\{\mathbf{x}_i'\}$, $\hat F$.
However, this is difficult to generalize to large dimensions.
Instead a probability density $\widehat P$ can be estimated from
the test data. It could be a parametric or
non-parametric estimation \cite[Chapter 7]{VapnikBook}, \cite{RissanenSpeedYu92}.
For a parametric estimation, $\widehat P_{\theta}$, we can choose the best hypothesis based on the
the (generalized) log-likelihood test (GLRT). We choose $H_0$ if
\[
\ell(\{\mathbf{x}_i'\})>\min_{\theta}-\sum_{i=1}^M\log_2 \widehat P_{\theta}(\mathbf{x}_i')+\tau
\]
otherwise, we choose $H_1$.
Here $\tau$ is the threshold in the hypothesis test that can
adjusted to obtain a desired false alarm probability. If hypothesis $H_1$ is chosen, then the test data, $\mathbf{x}'$, is considered to be from a different distribution than the training data $\mathbf{x}$. However, if a complicated
model $\widehat P_{\theta}$ with many parameters is used, it can easily
lead to overfitting, which results in a poor ROC (receiver operating
characteristic) curve. Our solution
is to instead consider a large class of alternative models
$\{\widehat P_\theta\}$, which includes both simple models and very complex models such
as non-parametric models like empirical pdfs \cite{RissanenSpeedYu92}. The complexity of the model is also accounted for through MDL principle
\cite{Rissanen83}. This leads to the MDL criterion where we choose $H_0$ if
\begin{align}
\ell(\{\mathbf{x}_i'\})>\min_{\widehat P_\theta}\min_{\theta}-\sum_{i=1}^M\log_2 \widehat P_{\theta}(\mathbf{x}_i')+L(\widehat P_\theta)+\tau. \label{eq:MDL}
\end{align}
Otherwise, we choose $H_1$. The term $L(\widehat P_\theta)$ is the description length of the model
$\widehat P_\theta$ within the class $\{\widehat P_\theta\}$. We can interpret the LHS and RHS of~\eqref{eq:MDL} as codelengths. This criterion is a special case of atypicality as defined in \cite{HostSabetiWalton15,SabetiHost17} 

\begin{definition}
\label{atypdef.thm}A test set, ${\mathbf{x}'_i}$ is (atypical) out of distribution if it can be described
(coded) with fewer bits in itself rather than using the (optimum)
code for (typical) training set, ${\mathbf{x}_i}$.
\end{definition}


A downside to using~\eqref{eq:MDL} is that it is  difficult to use the method on complex data which
do not have "nice" distributions. An equivalent, but more practical, approach to (atypical) out of distribution detection is
based on the observation that if data is encoded
by an optimum coder, the resulting bitstream
is iid uniform, i.e., totally random \cite{CoverBook}.
In other words, the output
of an optimum coder can not be compressed any further. If
it can be further compressed, it is therefore OOD.
From a theoretical point of view,
this approach is directly an application of the
theory of randomness developed by Kolmogorov
and Martin-L\"{o}f \cite{LiVitanyi,NiesBook,CoverBook}. 
Random sequences can be characterized through Kolmogorov
complexity. A sequence of bits $\{x_{n},n=1,\ldots,\infty\}$ is random
(i.e, iid uniform) if the Kolmogorov complexity of the sequence satisfies
$K(x_{1},\ldots,x_{n})\geq n-c$ for some constant $c$ and for all
$n$ \cite{LiVitanyi}. The sequence is incompressible if $K(x_{1},\ldots,x_{n}|n)\geq n$
for all $n$, and a finite sequence is algorithmically random if $K(x_{1},\ldots,x_{n}|n)\geq n$
\cite{CoverBook}. Thus, we characterize a set of data
as OOD if the output of the optimum
coder is not random according to 
Martin-L\"{o}f's definition.

For real-valued data, coding into a bitstream is not feasible
or useful. A better approach is to transform data into
an iid Gaussian random variables (i.e., $\mathcal{N}(0, \Sigma)$), and then use a coder for
the Gaussian distribution. The transformation is always
possible for continuous random variables \cite{SabetiHost17}
\begin{proposition}
\label{Generate.thm}For any continuous random variable $\mathbf{X}$
there exists an $n$-dimensional iid uniform random variable $\mathbf{U}$ and an invertible $\check{\mathbf{F}}$
so that $\mathbf{X}=\check{\mathbf{F}}^{-1}(\mathbf{U})$.
\end{proposition}
IID uniform random variables can be converted into an iid Gaussian distribution
by using the normal CDF $\Phi$ on each $U_i$. We will call
the combined invertible map $\mathcal{E}: \mathbb{R}^n\to \mathbb{R}^n$. Unlike~\eqref{eq:MDL}, this approach leads to an algorithm that can be applied to complex high-dimensional data: Algorithm \ref{alg:atypical}.

\begin{algorithm}[ht]
   \caption{Known $P(\mathbf{x})$}
   \label{alg:atypical}
\begin{algorithmic}[1]
   \State {\bfseries Input:} $\{\mathbf{x}'_i$, $i=1\ldots M\}$
\State Calculate $\mathbf{z}_i'=\mathcal{E}(\mathbf{x}_i)$.
\State Code the sequence $\{\mathbf{z}'_i$, $i=1\ldots M\}$
with an iid Gaussian coder with length $L_1$.
\State Code the sequence $\{\mathbf{z}'_i$, $i=1\ldots M\}$
with a universal vector Gaussian coder with length $L_2$.
   \If{$L_2+\tau < L_1$}
   \State $\{\mathbf{x}_i'$, $i=1\ldots M\}$ is OOD.
   \EndIf
\end{algorithmic}
\end{algorithm}
How to code multivariate Gaussian data will be discussed in
Section \ref{sec:coding}. A limitation is that the universal
coder (a coder that does not know the underlying data distribution apriori) is also Gaussian. This is because really the only "nice"
multivariate distribution is Gaussian. However, one could
easily combine this with a coder that codes each component
of $\mathbf{z}'_i$ with an MDL based histogram as in
\cite{RissanenSpeedYu92}; coders can always be combined by
choosing the shortest codelength as in (\ref{eq:MDL}). We did not pursue this direction because we think the most interesting signs of
distribution shift can be seen in the high dimensional
structure.

In practice, we generally do not know $P(\mathbf{x})$,
and in the following we will therefore generalize
Algorithm \ref{alg:atypical} to this case.

\subsection{$P(\mathbf{x})$  Unknown }
\label{sec:unknown}
When $P(\mathbf{x})$ is not known apriori, it first has to be estimated from a set of training data
$\mathbf{x}_i$, $i=1\ldots N$. Then to use
Algorithm \ref{alg:atypical} we need the (invertible) map $\mathcal{E}$. 
In one dimension one can simply use the empirical CDF, but
when the data is high-dimensional, we need a more sophisticated approach. In this paper we choose to use BiGAN \cite{donahue2016adversarial,dumoulin2016adversarially}. We will
discuss alternatives below.

\begin{wrapfigure}[11]{r}{0.55\textwidth}
    \centering
    \vspace{-0.2in}
    \includegraphics[width=80mm]{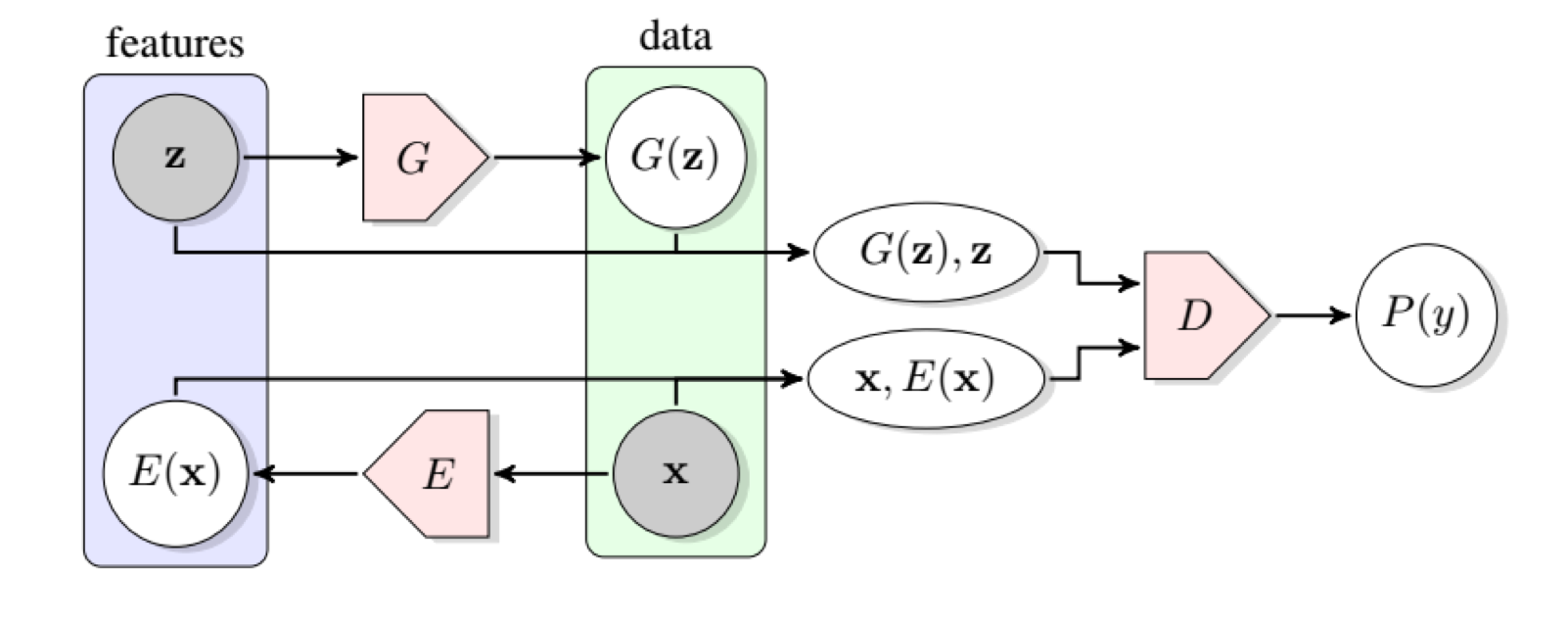}
    \caption{{ BiGAN structure from \cite{donahue2016adversarial} }}
    \label{fig:BiGANstructure}
\end{wrapfigure}


Shown in Figure~\ref{fig:BiGANstructure},  BiGAN has a generator map 
$\mathcal{G}:\mathbb{R}^m\to \mathbb{R}^n$ and an encoder
map $\mathcal{E}:\mathbb{R}^n\to \mathbb{R}^m$. Here
$\mathbb{R}^m$, $m\leq n$, is the feature or latent space, on which
the distribution is specified to be $\mathcal{N}(\mathbf{0}, \mathbf{I})$,
as is usual in GAN.
Extending the usual GAN architecture, the discriminator now
trains to jointly discriminate  in data and latent space (tuples $(\mathbf{x}, \mathcal{E}(\mathbf{x}))$ versus $(\mathcal{G}(\mathbf{z}), \mathbf{z}))$.


If the training of the BiGAN is perfect, the mappings
$\mathcal{G}$ and $\mathcal{E}$ are each others inverses according to \cite{donahue2016adversarial}.
However, in practical training scenarios, the solutions
for $\mathcal{G}$ and $\mathcal{E}$ do not necessarily converge to the
the global optimum. 
That means $\mathbf{x}\neq \mathcal{G}(\mathcal{E}(\mathbf{x}))$. Since MDL is based on lossless coding, that is,
reproducing data exactly from the encoding, this is an issue. 

Our solution is that we first encode
$\mathbf{z}=\mathcal{E}(\mathbf{x})$, as in in Algorithm
\ref{alg:atypical}, and subsequently also encode the residual,
or reconstruction error,
$\mathbf{x}-\mathcal{G}(\mathbf{z})$. We will detail how this
joint coding is done. Notice that MDL \cite{Rissanen83}
 is based on quantizing the real-valued data and
 then letting the quantization step converge to zero.
We need to carefully derive how this works for the joint
coding.

Consider coding of a \emph{single} data $\mathbf{x}$.
We represent each component of
$\mathbf{z}=\mathcal{E}(\mathbf{x})$ in
fixed point with
$r_z$ digits after '.' and unlimited number of digits
prior as in \cite{Rissanen83}. The number of bits required to encode
$\mathbf{z}$ is then
\begin{align}
L_z(\mathbf{z}) & =-\log_2\int_{\mathbf{z}}^{\mathbf{z}+2^{-r_z}}P_z(\mathbf{t})d\mathbf{t}\approx-\log_2(P_z(\mathbf{z})2^{-mr_z})\nonumber \\
 & =-\log_2 P_z(\mathbf{z})+mr_z.
\end{align}
Next the coder encodes $\mathbf{x}-\mathcal{G}(\hat{\mathbf{z}})$
with $r$ bits precision,
where $\hat{\mathbf{z}}$ is the quantized version
of $\mathcal{E}(\mathbf{x})$. The number of bits required is
\begin{align}
L(x) & =-\log_2\int_{\mathbf{x}}^{\mathbf{x}+2^{-r}}P(\mathbf{t}-\mathcal{G}(\hat{\mathbf{z}}))d\mathbf{t} \nonumber \\
& \approx-\log_2(P(\mathbf{x}-\mathcal{G}(\hat{\mathbf{z}}))2^{-nr})\nonumber \\
 & =-\log_2 P(\mathbf{x}-\mathcal{G}(\hat{\mathbf{z}}))+nr.
\end{align}
we have here assumed that $\hat{\mathbf{z}}$ does
not change inside the quantization cube
$[\mathbf{x},\mathbf{x}+2^{-r}]$. This is true if
the number of digits $r$ is large relative to $r_z$,
which is reasonable as even if let $r_z\to\infty$,
we do not get lossless coding due the non-invertibility
of $\mathcal{E}$, but
$r\to\infty$ gives lossless coding. Here
$mr_z$ and $nr$ do not depend on which coder is used,
and therefore cancels out: In the criterion (\ref{eq:MDL})
we always compare two codelengths and the
$mr_z + nr$ in both sides cancel out. We can therefore let
$r,r_z\to\infty$ and we end up with the codelength
\begin{align}
  -\log_2 P_z(\mathbf{z})-\log_2 P(\mathbf{x}-  \mathcal{G}(\mathbf{z}))
  \label{eq:combined}
\end{align}
as codelength $L_1$. The remaining question is
what to use for the distribution $ P(\mathbf{x}-\mathcal{G}({\mathbf{z}}))$. Since
the idea is that $\mathcal{E}$ should capture all structure in the data, we suggest to simply use an iid Gaussian distribution,
$\mathbf{x}-\mathcal{G}({\mathbf{z}})\sim \mathcal{N}(\mu \mathbf{1},\sigma^2\mathbf{I})$, with either estimated $\mu$,
$\sigma^2$ for training coding, or using
MDL for $\mu, \sigma^2$ in the universal coder. 

The method is described in Algorithm \ref{alg:unknown}.

\begin{algorithm}[ht]
   \caption{Unknown $P(\mathbf{x})$}
   \label{alg:unknown}
\begin{algorithmic}[1]
    \Statex {\textit{Training phase}}
   \State {\bfseries Input:} $\{\mathbf{x}_i$, $i=1\ldots N\}$
\State Train BiGAN on $\mathbf{x}_i$ with output
$\mathcal{G},\mathcal{E}$.
\State Estimate the covariance matrix of $\{\mathcal{E}(\mathbf{x}_i)$, $i=1\ldots N\}$ as described in Section \ref{sec:coding}.
\State Estimate the scalars $\mu$ and $\sigma^2$ of
$\{\mathbf{x}_i-\mathcal{G}(\mathcal{E}(\mathbf{x}_i))$, $i=1\ldots N\}$
\Statex{}
\Statex {\textit{Test phase}}
\State {\bfseries Input:} $\{\mathbf{x}'_i$, $i=1\ldots M\}$
\State Calculate $\mathbf{z}_i'=\mathcal{E}(\mathbf{x}_i)$.
\State Code the data $\{\mathbf{z}'_i$, $i=1\ldots M\}$
with the estimated covariance matrix from training using a multivariate Gaussian coder,
and $\{\mathbf{x}_i'-\mathcal{G}(\mathcal{E}(\mathbf{x}_i'))$, $i=1\ldots M\}$
with $\mu, \sigma^2$ from training using a univariate Guassian coder. Combine
using (\ref{eq:combined}) resulting in
 codelength $L_1$. 
\State Code the data $\{\mathbf{z}'_i$, $i=1\ldots M\}$
with a universal multivariate Gaussian coder as described in Section \ref{sec:coding}
and $\{\mathbf{x}_i'-\mathcal{G}(\mathcal{E}(\mathbf{x}_i'))$, $i=1\ldots M\}$
with a universal univariate Gaussian coder \cite[Section 13.2]{CoverBook}. Sum the two codelengths together to obtain $L_2$.

   \If{$L_2+\tau < L_1$}
   \State $\{\mathbf{x}_i'$, $i=1\ldots M\}$ is out-of-distribution.
   \EndIf
\end{algorithmic}
\end{algorithm}
The reason we code 
$\{\mathbf{z}_i'\}$ with an
estimated covariance matrix in
step 7 is that the distribution
of $\mathcal{E}(x)$ might not be perfectly $\mathcal{N}(\mathbf{0}, \mathbf{I})$, and
then the universal coder would always be better if we did not encode
with the estimated covariance matrix.

The main issue with BiGAN in our context is that 
the reconstruction error $\mathbf{x}-\mathcal{G}(\mathcal{E}(\mathbf{x}))$
can be large with an "ugly" distribution for both in-distribution
and OOD data. We did experiment with autoencoders as alternatives. A key
feature required of methods in our context is the ability
to prescribe the distribution in the latent space, and not
all autoencoders have that ability. We tried to alternative
methods that have that ability:
Adversarial AutoEncoder (AAE) \cite{makhzani2015adversarial}, and Variational AutoEncoder (VAE) \cite{kingma2013auto}. 
Both of these were better at reconstruction, i.e., $\mathbf{x}-\mathcal{G}(\mathcal{E}(\mathbf{x}))$ was significantly
smaller. However, the distribution of $\mathcal{E}(x)$ were
very far from Gaussian, where BiGAN gives something
that is close to Gaussian, and in the end BiGAN therefore had better performance.

\section{Universal Multivariate Gaussian Coder}
\label{sec:coding}
A multivariate Gaussian distribution with zero mean is completely characterized by its covariance matrix $\boldsymbol{\Sigma}$. When the covariance matrix is not known, it has to be estimated from data. A simple solution to estimate $\boldsymbol{\Sigma}$ is the maximum likelihood, which is empirical covariance matrix $\mathbf{S}$. However, the empirical covariance matrix is an overparametrized solution especially when the data is generated under a sparse Gaussian graphical model; that is $\boldsymbol{\Sigma}^{-1}$ is sparse \cite{Dempster72}. 
An efficient method for finding $\boldsymbol{\Sigma}^{-1}$, especially when a sparse structure is of interest, is to use regularized log-likelihood solution \cite{friedman2008sparse, yuan2007model}

\begin{equation}\label{eq.glasso}
   \underset{\boldsymbol{\Omega} \succ 0}{\text{max}} \hspace{0.2cm}  \mbox{logdet} \hspace{0.1cm} \boldsymbol{\Omega} -\mbox{tr}(\mathbf{S}\boldsymbol{\Omega}) - \lambda \|\boldsymbol{\Omega}\|_1,
\end{equation}
where $\boldsymbol{\Omega} = \boldsymbol{\Sigma}^{-1}$ is the precision matrix, $\lambda$ is the regularization term, and $\|\boldsymbol{\Omega}\|_1$ is $\ell_1$ norm. The parameter $\lambda$ determines the sparsity of $\boldsymbol{\Omega}$ and needs to be tuned. 
The sparsity pattern of $\boldsymbol{\Omega}$ can be captured by a graph called conditional independence graph $G(V, E)$, where $V$ is the set of vertices and $E$ is the set of edges. The graph $G$ contains an edge between vertex $i$ and vertex $j$, i.e., $(i,j) \in E$, if $\boldsymbol{\Omega}_{ij} \neq 0$.

To pick the best value of penalty term $\lambda$, different model selection techniques have been proposed in the literature. A detailed overview of existing methods can be found in \cite{Abolfazli21ISIT}. 
We use the approach proposed in \cite{abolfazli2021graph} to select $\lambda$ and hence find graph $G$. Unlike other approaches that may only consider the number of edges from $G$ (i.e., number of nonzero elements in $\boldsymbol{\Omega}$) to account for model complexity, this approach considers the whole structure of $G$ for a more accurate model complexity.
This is achieved by adopting MDL principle. For every value of $\lambda$ we first encode the structure associated with it, i.e., $G_\lambda$ and then we encode the data with the resulting sparse structure; the model that minimizes the summation of these two terms is the best model
\begin{equation}
    \min_{\lambda} L(G_{\lambda}) + L(\mathbf{x}|G_{\lambda}).
    \label{eq:MDL2}
\end{equation}

In \eqref{eq:MDL2}, $L(G_{\lambda})$ is the description length of the conditional independence graph, $G_{\lambda}$, associated with $\lambda$, and $L(\mathbf{x}|G_{\lambda})$ is the description length of data, $\mathbf{x}$, when encoded with $G_{\lambda}$. We use graph statistics such as graph motifs and degree distribution to encode $G_{\lambda}$. 
The detail on how to compute $L(G_{\lambda})$ can be found in \cite{abolfazli2021graph}.

To compute $L(\mathbf{x}|G_{\lambda})$, we need to encode the data with respect to $G_{\lambda}$. 
Based on the empirical covariance matrix $\mathbf{S}$, one can estimate the covariance matrix, $\widehat{\boldsymbol{\Sigma}}$, using Dempster's covariance selection \cite{Dempster72} that satisfies $\widehat{\boldsymbol{\Sigma}}_{ij} = \mathbf{S}_{ij}$ when $(i,j) \in E$, and $\widehat{\boldsymbol{\Sigma}}^{-1}_{ij} = 0$ when $(i,j) \notin E$.

Once we estimate $\widehat{\boldsymbol{\Sigma}}$, we use predictive MDL to compute $L(\mathbf{x}|G_{\lambda})$

\begin{equation}
    L(\mathbf{x}|G_{\lambda}) = -\sum_{i=0}^{N-1} \log_2 f\left(\mathbf{x}_{i+1}|\hat{\boldsymbol{\theta}}(\mathbf{x}_1,\ldots, \mathbf{x}_i)\right), \label{eq:predMDL}
\end{equation}

where $f(\cdot|\cdot)$ is the conditional pdf and $\hat{\boldsymbol{\theta}}(\mathbf{x}_1,\ldots, \mathbf{x}_i)$ denote the maximum likelihood estimate of parameters, which is $\widehat{\boldsymbol{\Sigma}}$ obtained under $G_{\lambda}$. 
To encode real-valued data, we assume a fixed-point representation with a (large) finite number, $r$, bits after the period, and an unlimited number of bits before the period as described in Section~\ref{sec:unknown}.
Note that since there is no estimate of $\widehat{\boldsymbol{\Sigma}}$ for the first few sample, we encode them with an arbitrary default distribution.

\section{Experiments}
\label{sec:experiment}
The proposed MDL method is examine with the digital image dataset MNIST \cite{deng2012mnist} as well as a real-world ECG dataset \cite{bratincsak2020electrocardiogram}. 

\subsection{MNIST dataset}
We used the \textit{ImageDataGenerator} function of the \textit{keras} Python library to perturb the MNIST dataset to create several synthetic OOD dataset. The corresponding perturbations to generate synthetic OOD datasets are listed in Table \ref{tab:scenarios}. 
\begin{table*}[htbp]
  \centering
  \caption{Scenarios for OOD detection. Rotation and shearing values are in degree, width and height shift in fraction, zoom and brightness in range.}
   \begin{tabular}{lcc}
        \toprule
        & Perturbation type & Value/Range value\\
        \hline
        \textsc{Case--1} & Rotation &  $5$ \\
        \textsc{Case--2}  & Shearing &   $20$\\
        \textsc{Case--3} & $[$ Width shift , Height shift$]$& $[0.02,0.02]$\\
        \textsc{Case--4} &Zooming& $[0.8,1.2]$ \\
        \textsc{Case--5} &Zooming& $[1,1.1]$ \\
        \textsc{Case--6} &Zooming& $[0.9,1]$ \\
        \textsc{Case--7} &Brightness & $[0.2,2]$ \\
        \textsc{Case--8} &Brightness & $[1,2]$ \\
        \textsc{Case--9} &Brightness & $[0.2,1]$ \\
        \textsc{Case--10} &Gaussian noise& $\mu = 0, \sigma = 0.05$ \\
        \bottomrule
        \label{tab:scenarios}
    \end{tabular}
\end{table*}

\begin{figure}[htbp]
\centering
\hspace{-0.3in}
\resizebox{\textwidth}{!}{\begin{tabular}{ccccc}
&
&
\subfloat[Original]{\includegraphics[width=1in]{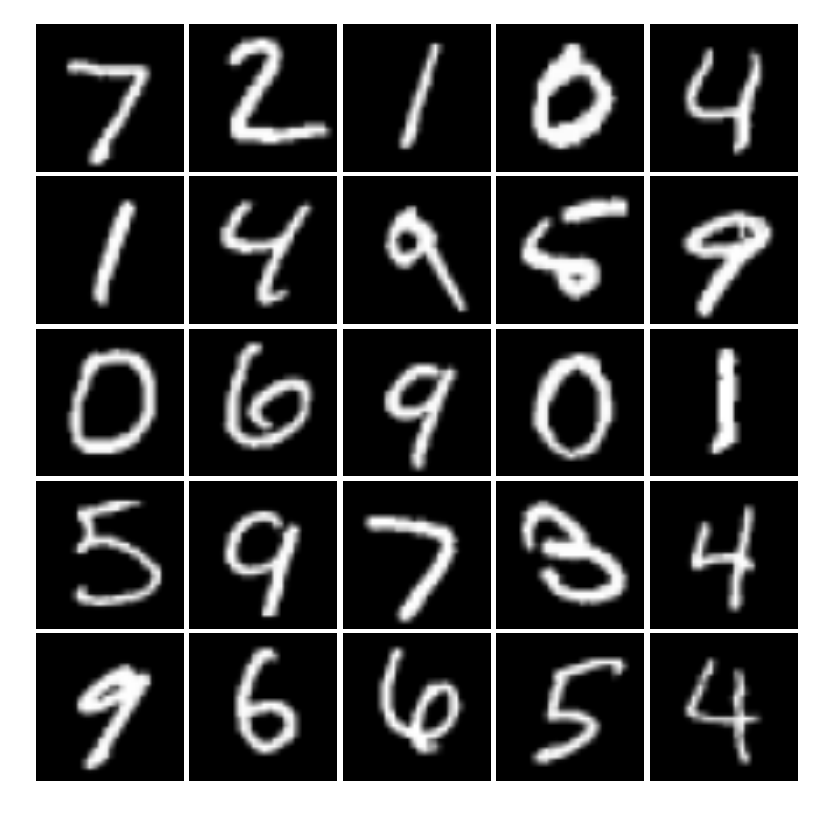}} \\ 
\subfloat[\textsc{Case--1}]{\includegraphics[width=1in]{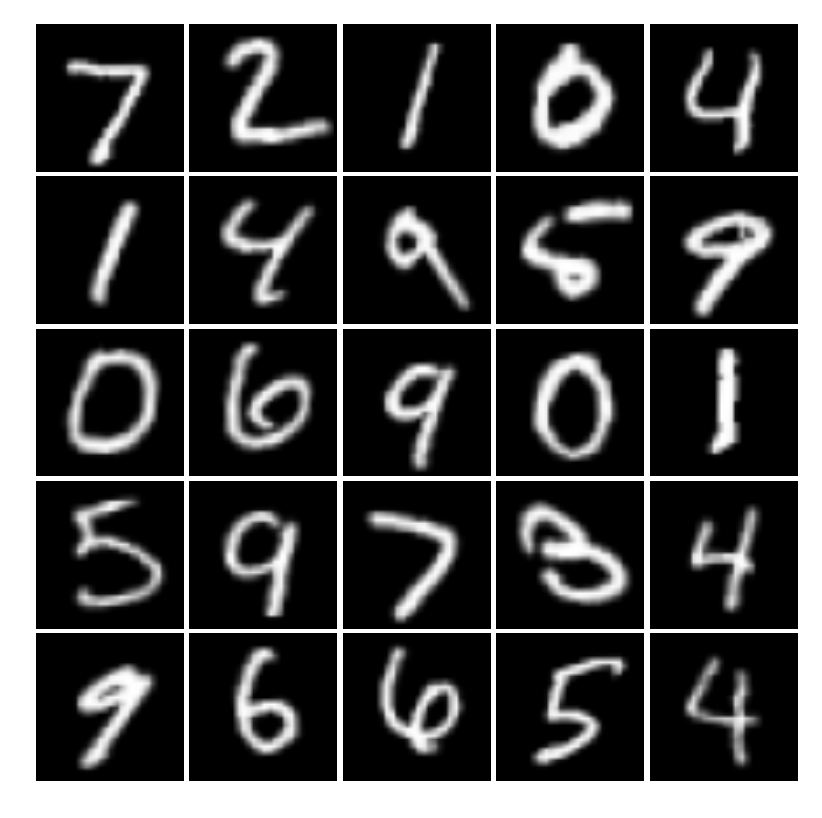}} & 
\subfloat[\textsc{Case--2}]{\includegraphics[width=1in]{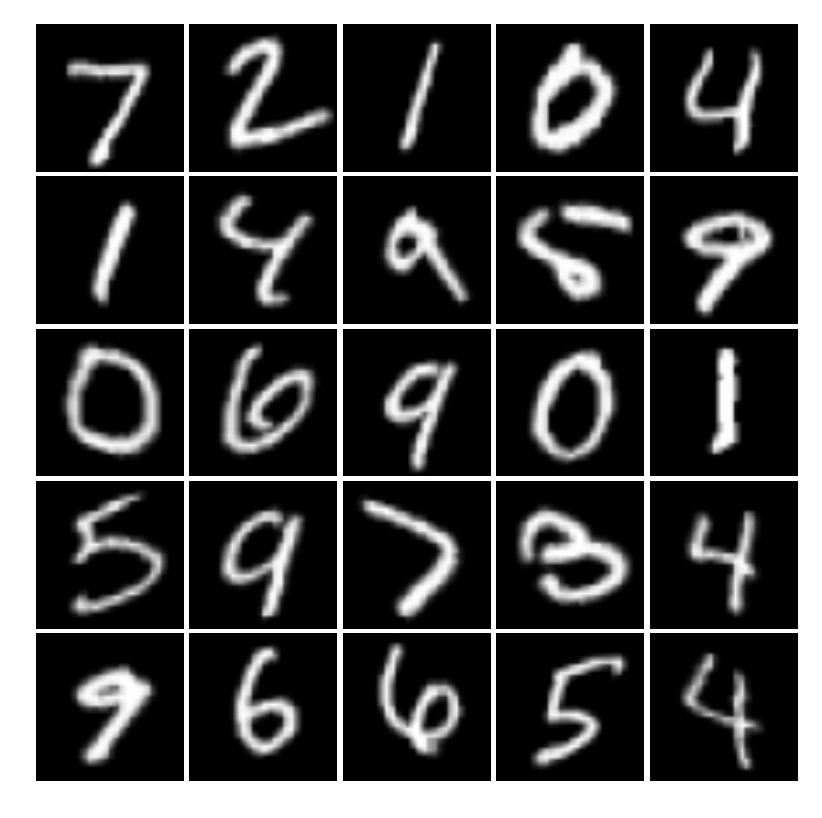}} & 
\subfloat[\textsc{Case--3}]{\includegraphics[width=1in]{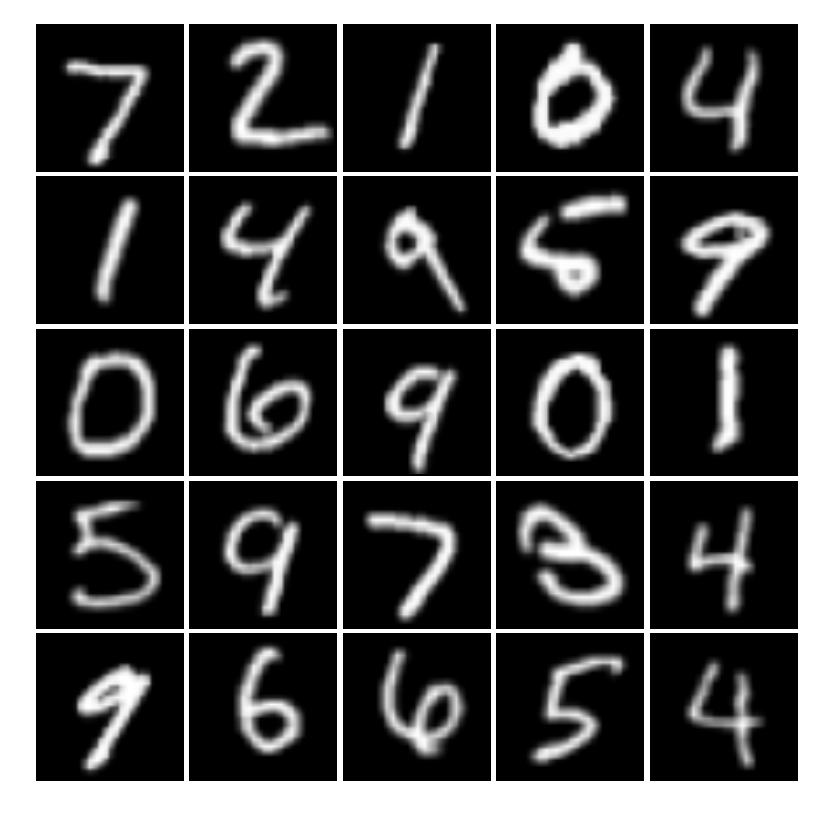}} & 
\subfloat[\textsc{Case--4}]{\includegraphics[width=1in]{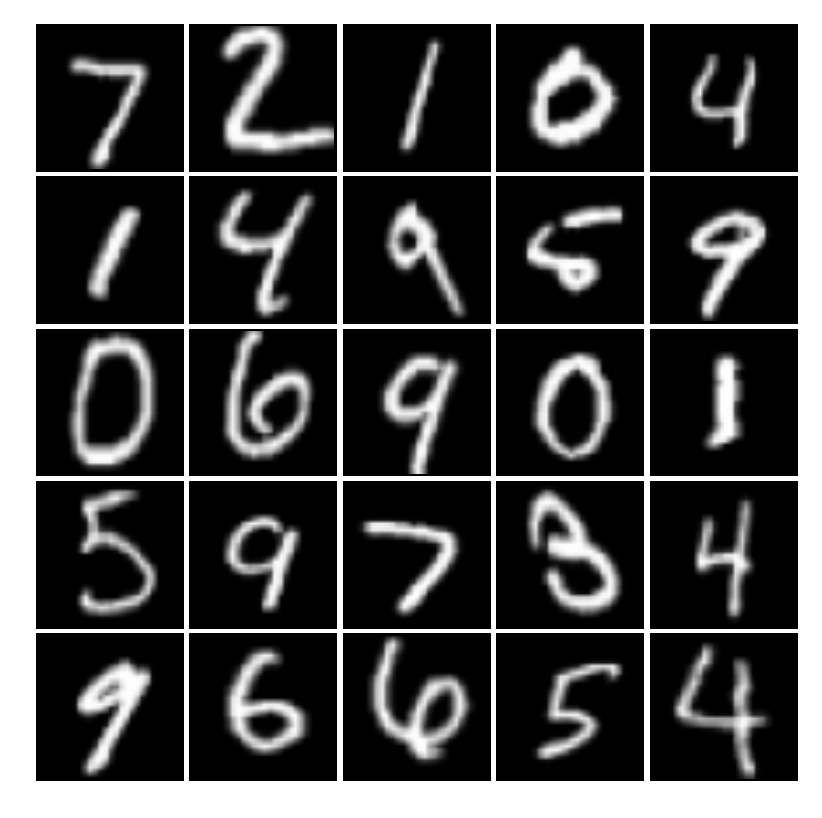}} & 
\subfloat[\textsc{Case--5}]{\includegraphics[width=1in]{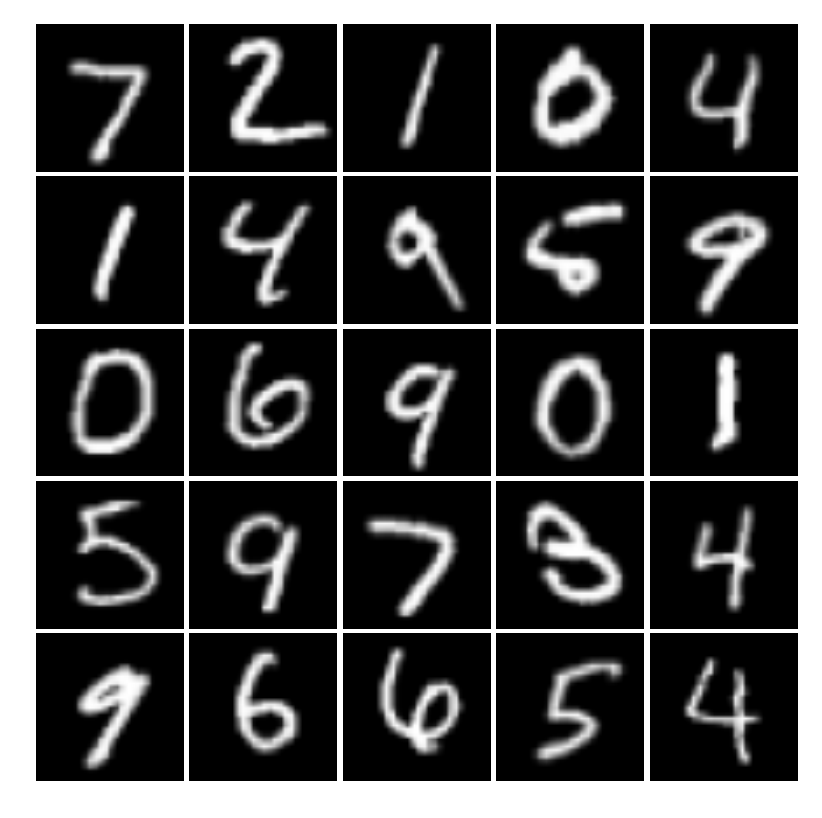}} \\
\subfloat[\textsc{Case--6}]{\includegraphics[width=1in]{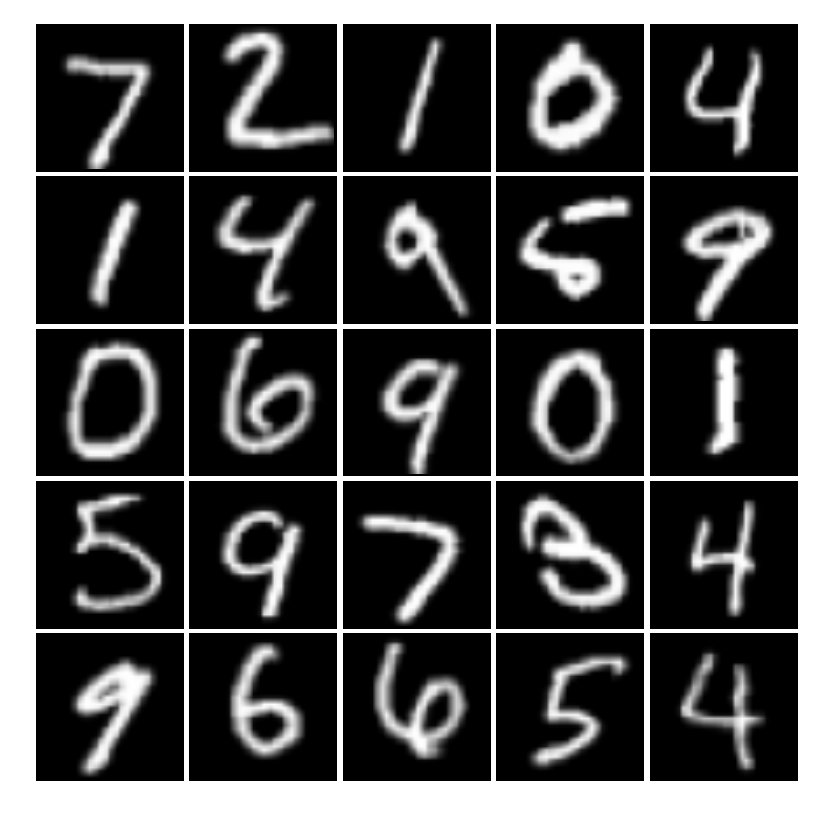}}&
\subfloat[\textsc{Case--7}]{\includegraphics[width=1in]{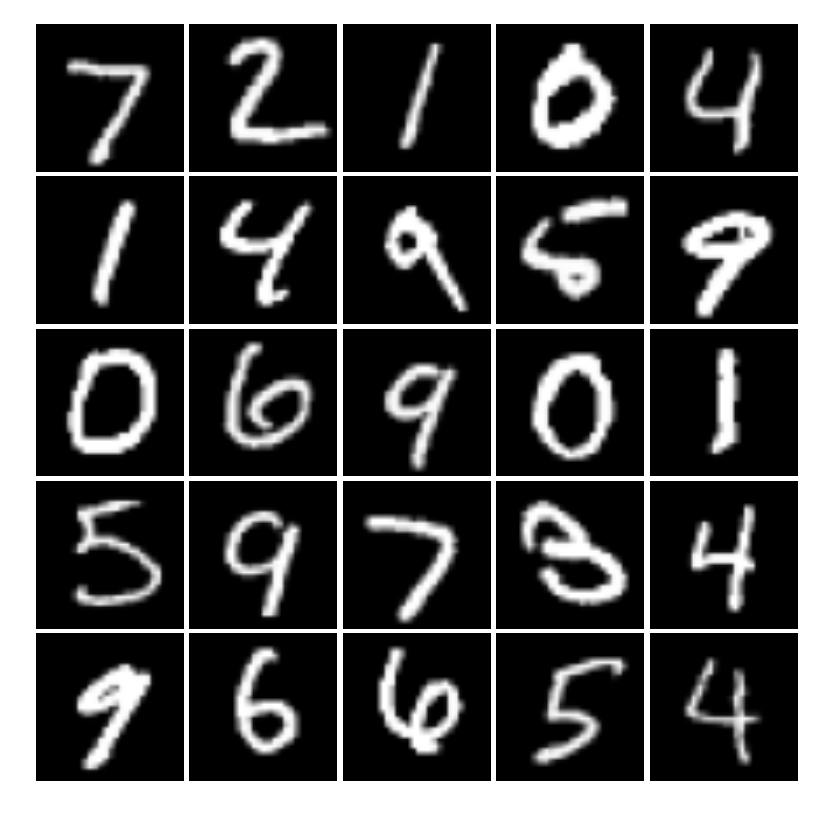}}&
\subfloat[\textsc{Case--8}]{\includegraphics[width=1in]{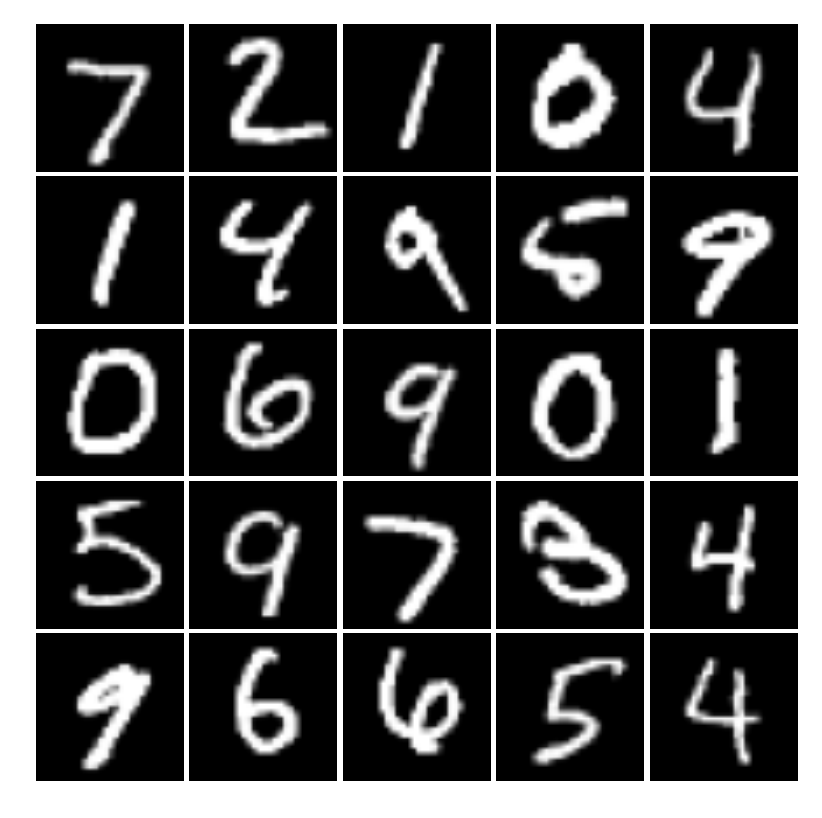}} & 
\subfloat[\textsc{Case--9}]{\includegraphics[width=1in]{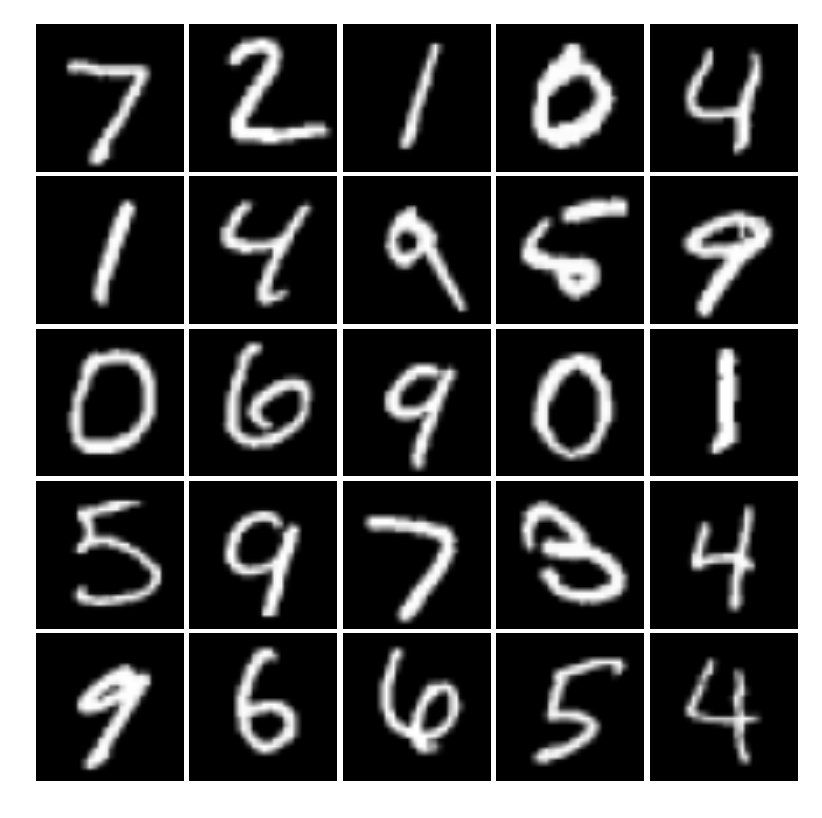}} &
\subfloat[\textsc{Case--10}]{\includegraphics[width=1in]{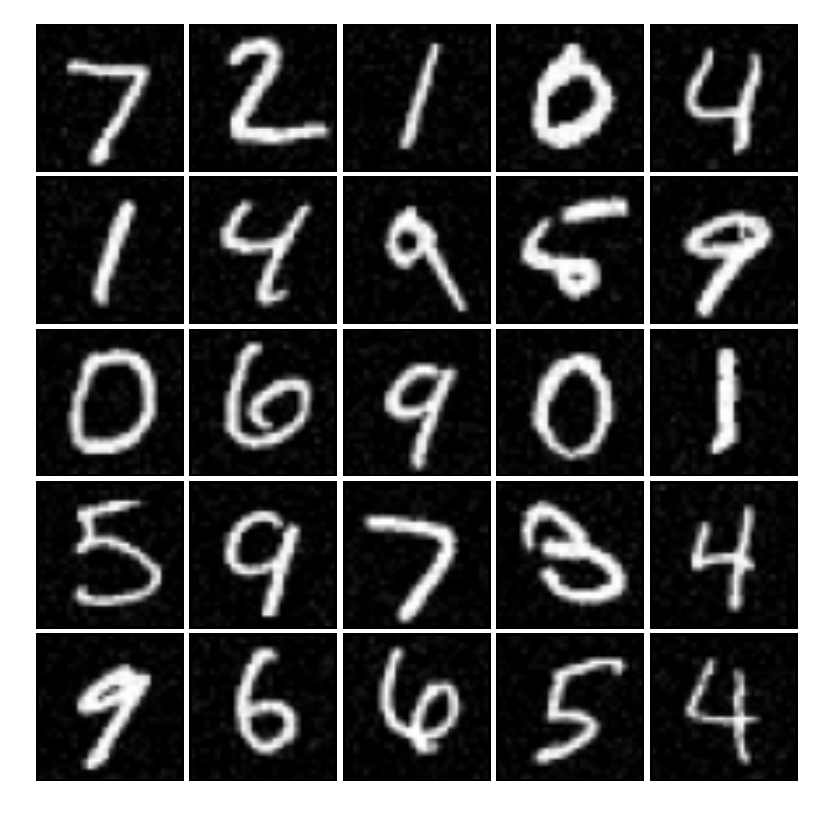}} \\
\end{tabular}}
\caption{Sample plot for original test images and corresponding images in OOD test cases. Note that the differences between the normal data and the OOD data is very subtle.}
\label{fig:Cases}
\end{figure}

As it can be seen in Figure~\ref{fig:Cases} it is hard to distinguish between in-distribution (unperturbed) and OOD (perturbed) data by visual inspection. In fact, each of samples in the OOD test set individually can be considered as a normal sample but their collective behavior is OOD. Our goal is exactly to detect such subtle changes.

In our experiment, 60,0000 examples in the original MNIST dataset were used for training and validation, and 10,000 original test images were used as in-distribution test data. All the modified datasets were considered as OOD in testing.

After training BiGAN, we transfer both training and test datasets into the latent space of size $20$. For the comparison scenarios, the AAE and VAE autoencoders as described in \cite{chalapathy2018group} were used to transfer the data into the latent space of the same dimension. Table \ref{tab:hyperparameters} provides the details of the employed BiGAN, AAE, and VAE  hyperparameters in this experiment. For the sake of fairness, we employ two fully connected hidden layers of size 512, each followed by a Leaky Relu and Batch normalization layers for the encoder and decode/generator networks in all three BiGAN, AAE, and VAE schemes. The same architecture is used for the discrimination network in BiGAN and AAE. The \textit{sigmoid} activation function in the last layer is used for the decoder/generator networks in all three schemes and in the discriminator network for BiGAN and AAE. 

\begin{table}[htbp]
    \caption{BiGAN, AAE, and VAE networks and training hyperparameters}
    \centering
    \begin{tabular}{lc}
    \toprule
Hyperparameter    & Value\\
\hline
        Number of epochs & $50,000$\\
         Batch size & $128$ \\
         Initial learning rate & $0.002$\\
         Learning rate decacy & $0.01$ after $25,000$ epochs \\
         $\ell_2$-regulizer & $2.5\cdot 10^{-4}$\\
         Initializing network weight & $\mathcal{N}(0,0.2)$\\
         Bath normalization momentum & $0.8$\\
         LeakyRelu & $\alpha = 0.2$\\
         \hline
    \end{tabular}
    \label{tab:hyperparameters}
\end{table}

Afterward, we encode the data using Algorithm~\ref{alg:unknown}, which entails the coding of data in latent space and encoding of reconstruction error.

We used graphical lasso \cite{friedman2008sparse} to solve \eqref{eq.glasso}. It is possible to use any regularized estimator of Gaussian graphical models. For regularization hyperparamter $\lambda$, we consider $20$ candidate values within the range $[0.1, 1]$ with logarithmic step size.
The results are given for one of the graph coders, i.e., coding with number of common neighbors and degree distribution. For details of coding, an interested reader can refer to \cite{abolfazli2021graph}.

To perform the experiments, we split the data into batches of size $M$ for each of the OOD test cases and also the original in-distribution test data and compute $L_2 - L_1$ based on \ref{alg:unknown} for each batch. 
Next, we can obtain ROC curve by plotting $L_2 - L_1$ of the OOD data versus $L_2 - L_1$ of the in-distribution test data. Table~\ref{tab:results} shows the area under ROC curve (AUROC) for each test case across different values of test set size $M$ for Algorithm~\ref{alg:unknown} verses the distance-based ($\ell_2$) approach proposed in \cite{chalapathy2018group}. We implemented the distance-based approach for three different generative networks: BiGAN, AAE, and VAE. The results indicate that our approach performed best in all cases except \textsc{Case--8}. Even for \textsc{Case--8}, our approach still had decent detection performance. The performance of Algorithm~\ref{alg:unknown} improves with test set size.

It might seem odd that  we get AUROC less than $0.5$ (i.e., random guessing) in some cases for $\ell_2$\textsc{-BiGAN}, $\ell_2$\textsc{-AAE}, and  $\ell_2$\textsc{-VAE}. The reason is that in these cases, the OOD data somehow has smaller reconstruction error than the in-distribution data even though the generative model has never seen the OOD data before. This phenomenon has also been observed in the literature \cite{choi2018waic,RabanserAl19,NalisnickAl19}.

Note that we can not improve the performance of the $\ell_2$-based approach by "flipping-the-output," i.e., decide the opposite of what the test says, since this does not apply to all the test cases such as \textsc{Case--2, 4, 8, 10}, and in real world we don't know what
is the type of OOD data.
\begin{table*}
\centering
\caption{AUROC for different test cases and testset size $M$. The best value for each case is boldfaced. Algorithm~\ref{alg:unknown} denotes our approach, and $\ell_2$ refer to method in \cite{chalapathy2018group} under different architectures.}
\resizebox{\textwidth}{!}{%
\begin{tabular}{l cccc| cccc | cccc }
\toprule
 & \multicolumn{4}{c}{$\mathbf{M=50}$} & \multicolumn{4}{c}{$\mathbf{M=100}$} & \multicolumn{4}{c}{$\mathbf{M=200}$}  \\
 & \begin{turn}{90}\textsc{Algorithm~\ref{alg:unknown}}\end{turn}  & \begin{turn}{90}$\ell_2$\textsc{-BiGAN} \end{turn}  & \begin{turn}{90}$\ell_2$\textsc{-AAE} \end{turn}& \begin{turn}{90}$\ell_2$\textsc{-VAE} \end{turn} &  \begin{turn}{90}\textsc{Algorithm~\ref{alg:unknown}}\end{turn} & \begin{turn}{90}$\ell_2$\textsc{-BiGAN} \end{turn} & \begin{turn}{90}$\ell_2$\textsc{-AAE} \end{turn} & \begin{turn}{90}$\ell_2$\textsc{-VAE} \end{turn}&  \begin{turn}{90}\textsc{Algorithm~\ref{alg:unknown}}\end{turn} & \begin{turn}{90}$\ell_2$\textsc{-BiGAN} \end{turn}  & \begin{turn}{90}$\ell_2$\textsc{-AAE} \end{turn} & \begin{turn}{90}$\ell_2$\textsc{-VAE} \end{turn} \\ 
\midrule
\textsc{Case--1}  &  $\mathbf{0.633}$   &     $0.167$   & $0.015$ & $0.003$ &   $\mathbf{0.693}$  &   $0.076$   &   $0.001$ &  $0.000$  &   $\mathbf{0.769}$    &   $0.032$ &  $0.000$ & $0.000$ \\
\textsc{Case--2}  &  $\mathbf{0.709}$   &    $0.685$   &  $0.505$ & $0.501$ &   $\mathbf{0.791}$   &   $0.765$   & $0.501$   &   $0.475$   &   $\mathbf{0.873}$     &   $0.839$   & $0.502$ & $0.511$\\
\textsc{Case--3}  &  $\mathbf{0.592}$   &    $0.192$   & $0.002$ & $0.000$ &    $\mathbf{0.638}$   &    $0.108$  & $0.000$ &   $0.000$   &   $\mathbf{0.687}$    &   $0.054$   & $0.000$   &   $0.000$\\
\textsc{Case--4}  &  $\mathbf{0.930}$   &    $0.928$   & $0.539$ & $0.464$ & $\mathbf{0.987}$   &  $0.980$    &  $0.606$ &   $0.503$ &  $\mathbf{1.000}$      &  $\mathbf{1.000}$    &    $0.573$ &    $0.477$\\
\textsc{Case--5}  &  $\mathbf{0.707}$   &    $0.176$   & $0.002$ & $0.001$ &  $\mathbf{0.815}$   &   $0.090$   & $0.000$ &   $0.000$ &  $\mathbf{0.914}$   &  $0.038$   & $0.000$   &   $0.000$ \\
\textsc{Case--6}  &  $\mathbf{0.840}$  & $0.645$   & $0.317$ &  $0.167$ &  $\mathbf{0.933}$  &   $0.711$   & $0.254$ &   $0.074$ & $\mathbf{0.989}$     &  $0.775$    & $0.174$ &   $0.021$\\
\textsc{Case--7}  &  $\mathbf{0.740}$   & $0.666$    &$0.139$ & $0.139$ &  $\mathbf{0.846}$   &   $0.722$   &    $0.050$ &   $0.067$ &  $\mathbf{0.917}$     &  $0.774$   &  $0.012$ &   $0.012$ \\
\textsc{Case--8}  &  $0.896$   & $0.903$    & $0.956$ & $\mathbf{0.982}$ &  $0.971$  &   $0.965$   &    $0.993$ &   $\mathbf{0.998}$ &  $0.997$    &  $0.997$    &   $\mathbf{1.000}$ &   $\mathbf{1.000}$  \\
\textsc{Case--9}  &  $\mathbf{0.988}$   & $0.197$    &  $0.000$   & $0.000$ &  $\mathbf{1.000}$   &   $0.111$  &   $0.000$   &   $0.000$   &  $\mathbf{1.000}$     &  $0.058$    &  $0.000$   &   $0.000$\\
\textsc{Case--10} &  $\mathbf{0.904}$   &    $0.712$ & $0.845$ & $0.663$ &  $\mathbf{0.967}$  &   $0.802$  & $0.923$ &   $0.725$ &  $\mathbf{0.998}$     &  $0.873$    & $0.978$ &   $0.807$\\
\bottomrule
\end{tabular}%
}
\label{tab:results}
\end{table*}

\subsection{ECG dataset}

We were provided ECG data of 23522 subjects, 423 subjects were diagnosed with the Kawasaki disease and the rest are considered  healthy subjects. Kawasaki disease causes inflammation in blood vessels in children under the age of five years old \cite{mccrindle2017diagnosis}. However, it can be very difficult for medical professionals to diagnose using since ECG data between healthy and diseased subjects can be very similar.

For each patient, a total of 176 Heart Rate Variation (HRV) features were automatically extracted from 12 leads ECG signals \cite{bratincsak2020electrocardiogram}. We used $90\%$ of the healthy (normal data) subjects were used for training. The remaining healthy subjects and diseased subjects were used as test set. A batch of healthy subjects should be detected as an in-distribution data whereas a batch of diseased subjects should be detected as an OOD sample. 

\begin{figure}[ht]
\vskip 0.2in
\begin{center}
\centerline{\includegraphics[width= 3.5in]{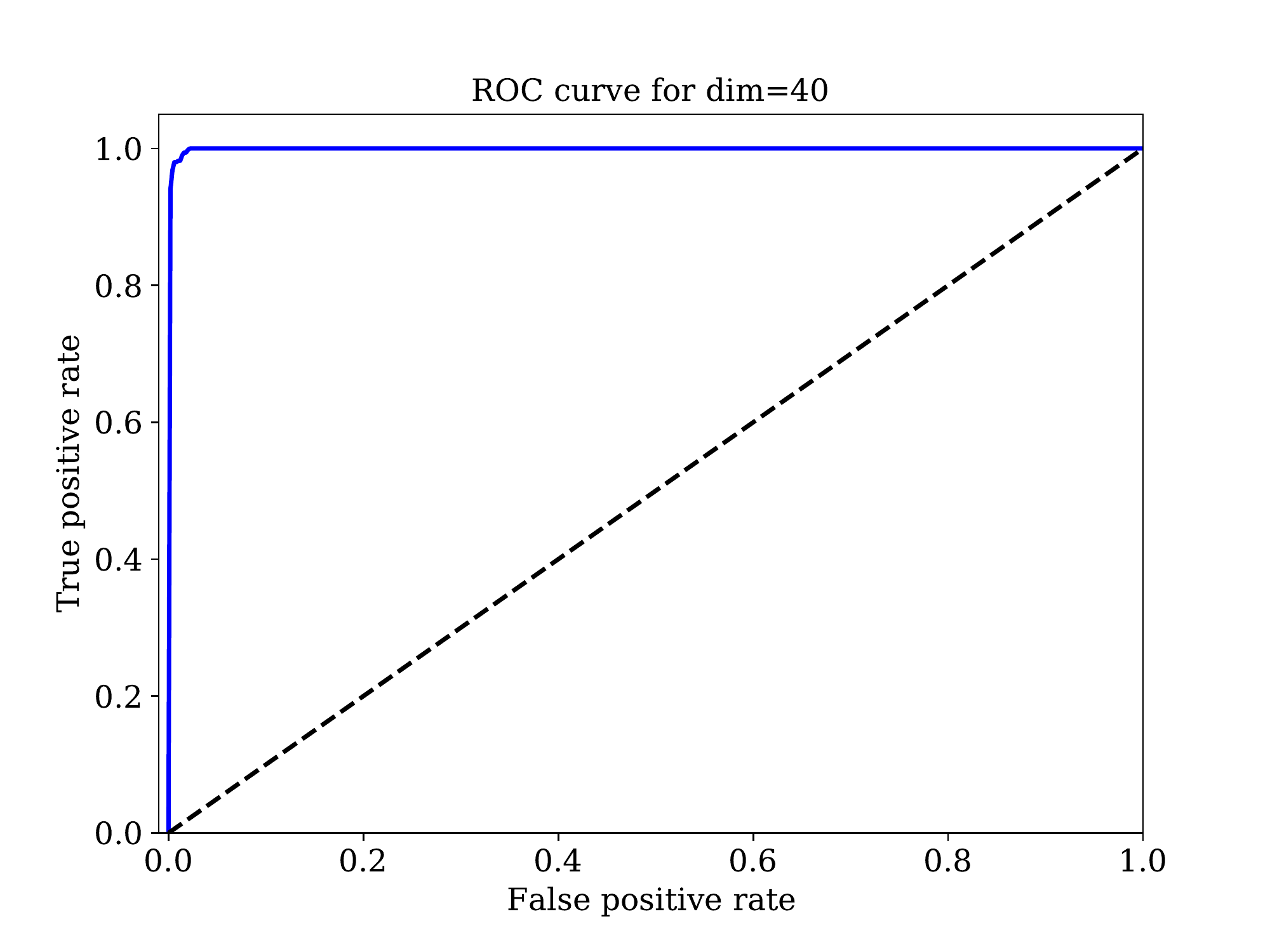}}
\caption{ROC curve of detecting Kawasaki subjects as OOD data.}
\label{roc:ECG}
\end{center}
\vskip -0.2in
\end{figure}

To test the performance of Algorithm~\ref{alg:unknown}, we randomly pick $50$ samples from each of in-distribution test data and OOD data (i.e., Kawasaki data). We repeat this experiment $500$ times. We used BiGAN, where the dimension of the latent space is $40$. The ROC curve is obtained by changing the threshold $\tau$. Figure~\ref{roc:ECG} represents ROC curve for this experiment. The corresponding AUROC is equal to $0.9996$. This shows that our approach can distinguish Kawasaki subjects as OOD data with high performance.

\section{Conclusion, Limitations, and
Future work}\label{sec:Conclusion}
We have developed a method for out-of-distribution
detection based on MDL. We have shown that
it performs well on both MNIST and real-world ECG data.

There are two future directions to consider. First, from our experiments, we learned that BiGAN is far
from ideal: the reconstruction error
$\mathbf{x}-\mathcal{G}(\mathcal{E}(\mathbf{x}))$
is quite large, and has a strange distribution. For future work, we will consider other deep generative models; we did experiment with AAE and VAE but they do resulted in additional issues compared with BiGAN. Second, we only considered multivariate Gaussian distributions as the alternative models in this paper. An advantage of our MDL/coding approach is that we can substitute a single coder with a combination of multiple coders. For future work, we will use such an approach to  consider alternative models that are mixtures of multivariate
Gaussians, and in more
generality non-parametric distributions.

\section*{Acknowledgements}
The research was funded in part by the NSF grant CCF-1908957.

\bibliography{Coop06,ahmref2,Coop03,BigData,reference}

\begin{thebibliography}{33}
\providecommand{\natexlab}[1]{#1}
\providecommand{\url}[1]{\texttt{#1}}
\expandafter\ifx\csname urlstyle\endcsname\relax
  \providecommand{\doi}[1]{doi: #1}\else
  \providecommand{\doi}{doi: \begingroup \urlstyle{rm}\Url}\fi

\bibitem[Abolfazli et~al.(2021{\natexlab{a}})Abolfazli, H{\o}st-Madsen, Zhang,
  and Bratincsak]{Abolfazli21ISIT}
Abolfazli, M., H{\o}st-Madsen, A., Zhang, J., and Bratincsak, A.
\newblock Graph coding for model selection and anomaly detection in gaussian
  graphical models.
\newblock In \emph{ISIT'2021, Melbourne, Australia, July 12-20, 2021},
  2021{\natexlab{a}}.

\bibitem[Abolfazli et~al.(2021{\natexlab{b}})Abolfazli, Host-Madsen, Zhang, and
  Bratincsak]{abolfazli2021graph}
Abolfazli, M., Host-Madsen, A., Zhang, J., and Bratincsak, A.
\newblock Graph compression with application to model selection.
\newblock \emph{arXiv preprint arXiv:2110.00701}, 2021{\natexlab{b}}.

\bibitem[Bergamin et~al.(2022)Bergamin, Mattei, Havtorn, Senetaire, Schmutz,
  Maal{\o}e, Hauberg, and Frellsen]{bergamin2022model}
Bergamin, F., Mattei, P.-A., Havtorn, J.~D., Senetaire, H., Schmutz, H.,
  Maal{\o}e, L., Hauberg, S., and Frellsen, J.
\newblock Model-agnostic out-of-distribution detection using combined
  statistical tests.
\newblock In \emph{International Conference on Artificial Intelligence and
  Statistics}, pp.\  10753--10776. PMLR, 2022.

\bibitem[Bishop(1994)]{bishop1994novelty}
Bishop, C.~M.
\newblock Novelty detection and neural network validation.
\newblock \emph{IEE Proceedings-Vision, Image and Signal processing},
  141\penalty0 (4):\penalty0 217--222, 1994.

\bibitem[Bratincs{\'a}k et~al.(2020)Bratincs{\'a}k, Kimata, Limm-Chan, Vincent,
  Williams, and Perry]{bratincsak2020electrocardiogram}
Bratincs{\'a}k, A., Kimata, C., Limm-Chan, B.~N., Vincent, K.~P., Williams,
  M.~R., and Perry, J.~C.
\newblock {Electrocardiogram standards for children and young adults using
  Z-scores}.
\newblock \emph{Circulation: Arrhythmia and Electrophysiology}, 13\penalty0
  (8):\penalty0 e008253, 2020.

\bibitem[Chalapathy et~al.(2018)Chalapathy, Toth, and
  Chawla]{chalapathy2018group}
Chalapathy, R., Toth, E., and Chawla, S.
\newblock Group anomaly detection using deep generative models.
\newblock In \emph{Joint European Conference on Machine Learning and Knowledge
  Discovery in Databases}, pp.\  173--189. Springer, 2018.

\bibitem[Choi et~al.(2018)Choi, Jang, and Alemi]{choi2018waic}
Choi, H., Jang, E., and Alemi, A.~A.
\newblock Waic, but why? generative ensembles for robust anomaly detection.
\newblock \emph{arXiv preprint arXiv:1810.01392}, 2018.

\bibitem[Corder \& Foreman(2014)Corder and Foreman]{corder2014nonparametric}
Corder, G.~W. and Foreman, D.~I.
\newblock \emph{{Nonparametric Statistics: A Step-by-Step Approach}}.
\newblock John Wiley \& Sons, 2014.

\bibitem[Cover \& Thomas(2006)Cover and Thomas]{CoverBook}
Cover, T. and Thomas, J.
\newblock \emph{{Information Theory, 2nd Edition}}.
\newblock John Wiley, 2006.

\bibitem[Dempster(1972)]{Dempster72}
Dempster, A.~P.
\newblock Covariance selection.
\newblock \emph{Biometrics}, 28\penalty0 (1):\penalty0 157--175, 1972.
\newblock ISSN 0006341X, 15410420.
\newblock URL \url{http://www.jstor.org/stable/2528966}.

\bibitem[Deng(2012)]{deng2012mnist}
Deng, L.
\newblock The mnist database of handwritten digit images for machine learning
  research.
\newblock \emph{IEEE Signal Processing Magazine}, 29\penalty0 (6):\penalty0
  141--142, 2012.

\bibitem[Donahue et~al.(2016)Donahue, Kr{\"a}henb{\"u}hl, and
  Darrell]{donahue2016adversarial}
Donahue, J., Kr{\"a}henb{\"u}hl, P., and Darrell, T.
\newblock Adversarial feature learning.
\newblock \emph{arXiv preprint arXiv:1605.09782}, 2016.

\bibitem[Dumoulin et~al.(2016)Dumoulin, Belghazi, Poole, Mastropietro, Lamb,
  Arjovsky, and Courville]{dumoulin2016adversarially}
Dumoulin, V., Belghazi, I., Poole, B., Mastropietro, O., Lamb, A., Arjovsky,
  M., and Courville, A.
\newblock Adversarially learned inference.
\newblock \emph{arXiv preprint arXiv:1606.00704}, 2016.

\bibitem[Friedman et~al.(2008)Friedman, Hastie, and
  Tibshirani]{friedman2008sparse}
Friedman, J., Hastie, T., and Tibshirani, R.
\newblock Sparse inverse covariance estimation with the graphical lasso.
\newblock \emph{Biostatistics}, 9\penalty0 (3):\penalty0 432--441, 2008.

\bibitem[H{\o}st-Madsen et~al.(2019)H{\o}st-Madsen, Sabeti, and
  Walton]{HostSabetiWalton15}
H{\o}st-Madsen, A., Sabeti, E., and Walton, C.
\newblock Data discovery and anomaly detection using atypicality.
\newblock \emph{IEEE Transactions on Information Theory}, 65\penalty0 (9),
  September 2019.

\bibitem[Kingma \& Welling(2013)Kingma and Welling]{kingma2013auto}
Kingma, D.~P. and Welling, M.
\newblock Auto-encoding variational bayes.
\newblock \emph{arXiv preprint arXiv:1312.6114}, 2013.

\bibitem[Li \& Vit\'anyi(2008)Li and Vit\'anyi]{LiVitanyi}
Li, M. and Vit\'anyi, P.
\newblock \emph{An Introduction to Kolmogorov Complexity and Its Applications}.
\newblock Springer, third edition, 2008.

\bibitem[Makhzani et~al.(2015)Makhzani, Shlens, Jaitly, Goodfellow, and
  Frey]{makhzani2015adversarial}
Makhzani, A., Shlens, J., Jaitly, N., Goodfellow, I., and Frey, B.
\newblock Adversarial autoencoders.
\newblock \emph{arXiv preprint arXiv:1511.05644}, 2015.

\bibitem[McCrindle et~al.(2017)McCrindle, Rowley, Newburger, Burns, Bolger,
  Gewitz, Baker, Jackson, Takahashi, Shah, et~al.]{mccrindle2017diagnosis}
McCrindle, B.~W., Rowley, A.~H., Newburger, J.~W., Burns, J.~C., Bolger, A.~F.,
  Gewitz, M., Baker, A.~L., Jackson, M.~A., Takahashi, M., Shah, P.~B., et~al.
\newblock Diagnosis, treatment, and long-term management of kawasaki disease: a
  scientific statement for health professionals from the american heart
  association.
\newblock \emph{Circulation}, 135\penalty0 (17):\penalty0 e927--e999, 2017.

\bibitem[Morningstar et~al.(2021)Morningstar, Ham, Gallagher, Lakshminarayanan,
  Alemi, and Dillon]{morningstar2021density}
Morningstar, W., Ham, C., Gallagher, A., Lakshminarayanan, B., Alemi, A., and
  Dillon, J.
\newblock Density of states estimation for out of distribution detection.
\newblock In \emph{International Conference on Artificial Intelligence and
  Statistics}, pp.\  3232--3240. PMLR, 2021.

\bibitem[Nalisnick et~al.(2018)Nalisnick, Matsukawa, Teh, Gorur, and
  Lakshminarayanan]{nalisnick2018deep}
Nalisnick, E., Matsukawa, A., Teh, Y.~W., Gorur, D., and Lakshminarayanan, B.
\newblock Do deep generative models know what they don't know?
\newblock \emph{arXiv preprint arXiv:1810.09136}, 2018.

\bibitem[Nalisnick et~al.(2019)Nalisnick, Matsukawa, Teh, and
  Lakshminarayanan]{NalisnickAl19}
Nalisnick, E., Matsukawa, A., Teh, Y.~W., and Lakshminarayanan, B.
\newblock Detecting out-of-distribution inputs to deep generative models using
  typicality, 2019.
\newblock URL \url{https://arxiv.org/abs/1906.02994}.

\bibitem[Nies(2009)]{NiesBook}
Nies, A.
\newblock \emph{Computability and Randomness}.
\newblock Oxford University Press, 2009.

\bibitem[Pimentel et~al.(2014)Pimentel, Clifton, Clifton, and
  Tarassenko]{Pimentel14}
Pimentel, M.~A., Clifton, D.~A., Clifton, L., and Tarassenko, L.
\newblock A review of novelty detection.
\newblock \emph{Signal Processing}, 99:\penalty0 215--249, 2014.
\newblock ISSN 0165-1684.
\newblock \doi{https://doi.org/10.1016/j.sigpro.2013.12.026}.
\newblock URL
  \url{https://www.sciencedirect.com/science/article/pii/S016516841300515X}.

\bibitem[Rabanser et~al.(2019)Rabanser, G\"{u}nnemann, and
  Lipton]{RabanserAl19}
Rabanser, S., G\"{u}nnemann, S., and Lipton, Z.
\newblock Failing loudly: An empirical study of methods for detecting dataset
  shift.
\newblock In \emph{Advances in Neural Information Processing Systems},
  volume~32. Curran Associates, Inc., 2019.
\newblock URL
  \url{https://proceedings.neurips.cc/paper/2019/file/846c260d715e5b854ffad5f70a516c88-Paper.pdf}.

\bibitem[Rao(1948)]{rao1948large}
Rao, C.~R.
\newblock Large sample tests of statistical hypotheses concerning several
  parameters with applications to problems of estimation.
\newblock In \emph{Mathematical Proceedings of the Cambridge Philosophical
  Society}, volume~44, pp.\  50--57. Cambridge University Press, 1948.

\bibitem[Rissanen(1978)]{Rissanen78}
Rissanen, J.
\newblock Modeling by shortest data description.
\newblock \emph{Automatica}, pp.\  465--471, 1978.

\bibitem[Rissanen(1983)]{Rissanen83}
Rissanen, J.
\newblock A universal prior for integers and estimation by minimum description
  length.
\newblock \emph{The Annals of Statistics}, \penalty0 (2):\penalty0 416--431,
  1983.

\bibitem[Rissanen et~al.(1992)Rissanen, Speed, and Yu]{RissanenSpeedYu92}
Rissanen, J., Speed, T., and Yu, B.
\newblock Density estimation by stochastic complexity.
\newblock \emph{IEEE Transactions on Information Theory}, 38\penalty0
  (2):\penalty0 315--323, 1992.
\newblock \doi{10.1109/18.119689}.

\bibitem[Sabeti \& H{\o}st-Madsen(2019)Sabeti and H{\o}st-Madsen]{SabetiHost17}
Sabeti, E. and H{\o}st-Madsen, A.
\newblock Data discovery and anomaly detection using atypicality for
  real-valued data.
\newblock \emph{Entropy}, pp.\  219, Feb. 2019.
\newblock Available at {https://doi.org/10.3390/e21030219}.

\bibitem[Serr{\`a} et~al.(2019)Serr{\`a}, {\'A}lvarez, G{\'o}mez, Slizovskaia,
  N{\'u}{\~n}ez, and Luque]{serra2019input}
Serr{\`a}, J., {\'A}lvarez, D., G{\'o}mez, V., Slizovskaia, O., N{\'u}{\~n}ez,
  J.~F., and Luque, J.
\newblock Input complexity and out-of-distribution detection with
  likelihood-based generative models.
\newblock In \emph{International Conference on Learning Representations}, 2019.

\bibitem[Vapnik(1998)]{VapnikBook}
Vapnik, V.~N.
\newblock \emph{Statistical Learning Theory}.
\newblock John Wiley, 1998.

\bibitem[Yuan \& Lin(2007)Yuan and Lin]{yuan2007model}
Yuan, M. and Lin, Y.
\newblock Model selection and estimation in the gaussian graphical model.
\newblock \emph{Biometrika}, 94\penalty0 (1):\penalty0 19--35, 2007.

\end{thebibliography}
\bibliographystyle{icml2022}


\end{document}